\documentclass[letterpaper]{article} 
\usepackage{aaai24}  
\usepackage{times}  
\usepackage{helvet}  
\usepackage{courier}  
\usepackage[hyphens]{url}  
\usepackage{graphicx} 
\usepackage{natbib}  
\usepackage{caption} 
\usepackage{algorithm}
\usepackage{algorithmic}
\usepackage{color}
\usepackage{amsfonts}

\usepackage{newfloat}
\usepackage{listings}
\DeclareCaptionStyle{ruled}{labelfont=normalfont,labelsep=colon,strut=off} 
\floatstyle{ruled}
\newfloat{listing}{tb}{lst}{}
\floatname{listing}{Listing}
\usepackage{booktabs}
\usepackage{relsize}
\usepackage{marvosym}
\usepackage{ifsym}

\title{\textsc{Prefer}: Prompt Ensemble Learning via Feedback-Reflect-Refine}
\author{
    Chenrui Zhang\textsuperscript{\rm 1\Letter},
    Lin Liu\textsuperscript{\rm 2}\footnote{This work was done during the internship at Meituan.},
    Jinpeng Wang\textsuperscript{\rm 1}, \\
    Chuyuan Wang\textsuperscript{\rm 1},
    Xiao Sun\textsuperscript{\rm 1},
    Hongyu Wang\textsuperscript{\rm 1},
    Mingchen Cai\textsuperscript{\rm 1}
}
\affiliations{
    \textsuperscript{\rm 1}Meituan Inc., Beijing, China
    \textsuperscript{\rm 2}Beijing Jiaotong University, Beijing, China\\
    \textsuperscript{\Letter}chenrui.zhang@pku.edu.cn, linliu@bjtu.edu.cn, \{wangjinpeng04,wangchuyuan,
sunxiao10,wanghongyu15,caimingchen\}@meituan.com

}

\usepackage{bibentry}

\begin{document}

\maketitle

\begin{abstract}
As an effective tool for eliciting the power of Large Language Models (LLMs), prompting has recently demonstrated unprecedented abilities across a variety of complex tasks. To further improve the performance, prompt ensemble has attracted substantial interest for tackling the hallucination and instability of LLMs.
However, existing methods usually adopt a two-stage paradigm, which requires a pre-prepared set of prompts with substantial manual effort, and is unable to perform directed optimization for different weak learners.
In this paper, we propose a simple, universal, and automatic method named \textsc{Prefer} (\textbf{P{\smaller R}}ompt \textbf{E}nsemble learning via \textbf{F}eedback-R\textbf{\smaller E}flect-\textbf{R}efine) to address the stated limitations.
Specifically, given the fact that weak learners are supposed to focus on hard examples during boosting, \textsc{Prefer} builds a feedback mechanism for reflecting on the inadequacies of existing weak learners. Based on this, the LLM is required to automatically synthesize new prompts for iterative refinement.
Moreover, to enhance stability of the prompt effect evaluation, we propose a novel prompt bagging method involving forward and backward thinking, which is superior to majority voting and is beneficial for both feedback and weight calculation in boosting.
Extensive experiments demonstrate that our \textsc{Prefer} achieves state-of-the-art performance in multiple types of tasks by a significant margin. We have made our code publicly available\footnote{https://github.com/zcrwind/PREFER}.


\end{abstract}

\vspace{-0.19cm}
\section{Introduction}
Large Language Models (LLMs) have recently flourished across a variety of fields, demonstrating unprecedented abilities in myriad of complex tasks~\cite{zhao2023survey,ouyang2022training}. Trained with large-scale web data on massive parameters, LLMs show emergent abilities beyond the original linguistic competence~\cite{wei2022emergent}, which perform tremendous versatility in both academia and industry. 
To elicit the power of pretrained LLMs directly or adapt LLMs to specific domains, various paradigms are proposed, including prompt engineering~\cite{qiao2022reasoning}, p-tuning~\cite{liu2021gpt}, and LoRA finetuning~\cite{hu2021lora}, etc.
Due to the immense scale of the model parameters, finetuning on all or even part of LLMs is costly and time-consuming. To this end, as a simple and effective paradigm, prompt engineering explores a fundamentally new way of invoking intrinsic knowledge and reasoning ability of LLMs based on a pretrain-prompt-predict manner~\cite{liu2023pre}.

\begin{figure}[t]
\setlength{\abovecaptionskip}{-1pt} 
\setlength{\belowcaptionskip}{-7pt}
\centering
\includegraphics[width=0.48\textwidth]{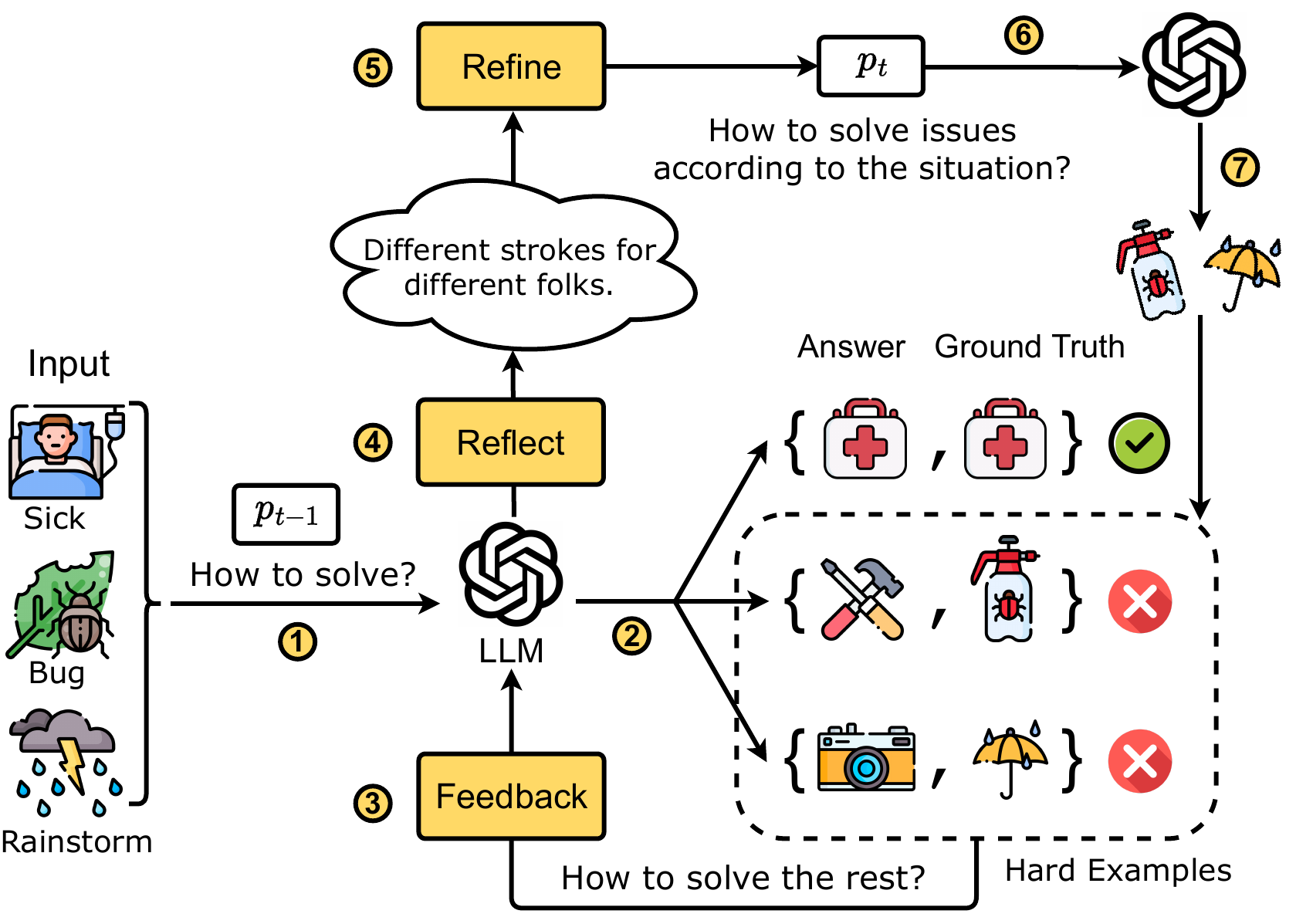} 
\vspace{-0.1cm}
\caption{High-level overview of feedback-reflect-refine paradigm. $p_t$ denotes the prompt at the $t$-th iteration.}
\label{fig:main_idea}
\vspace{-0.1cm}
\end{figure}
Though promising, the naïve prompting approaches are afflicted by several limitations. 
As generative language models, LLMs' output commonly has a large variance. For instance, the reasoning logic and predicted results could be contradictory in multiple runs, although the input prompts are fixed.
In addition, LLMs suffer from the notoriously hallucination issue~\cite{ji2023survey}, leading to results that are plausible-sounding but factually incorrect or irrelevant to the inputs. 
Furthermore, the quality of LLMs’ output is susceptible to the given prompts, which entails substantial manual effort and domain expertise to find out the reliable prompts.

As a promising solution to these issues, prompt ensemble learning has attracted substantial interest in the community very recently, demonstrating significant improvements in both effectiveness and stability across various tasks.
As a representative work, PromptBoosting~\cite{hou2023promptboosting} applies the traditional \textsc{Ada}\textsc{Boost}~\cite{freund1997decision} algorithm over a set of pre-defined prompts for text classification.
BPE~\cite{pitis2023boosted} focuses on Chain-of-Thought (CoT)~\cite{wei2022chain} boosting and builds few-shot CoT prompts based on self-consistency~\cite{wang2022self}.
These efforts empirically demonstrate the strength of prompt ensembles for LLM-based tasks, yielding exceptional performance gains over single-prompt baselines.


However, despite their success, existing prompt ensemble approaches, which typically adopt a two-stage process, have several limitations. First, they require a pre-prepared set of prompts in advance, which are either manually defined or generated by another language model with heavy parameters. This preliminary work is costly and laborious, often involving a trial-and-error or pre-evaluation process to ensure the quality of pre-defined prompts. Second, the two-stage paradigm fixes the prompts to be used in the ensemble process, limiting the adaptability and scalability of prompt boosting, as the prompts cannot be optimized jointly. Since the relationships between prompts are ignored during the iterative boosting process, the pre-defined prompts tend to be sub-optimal and susceptible. Moreover, existing methods conduct ensembles either in boosting or in bagging individually, neglecting the potential benefits of combining the two worlds to enhance performance.

To alleviate the above issues, we advocate that a smarter paradigm for prompt ensemble in the era of LLMs is expected to be automatic, self-adaptive and joint-optimizable. Such paradigm reduces the need for manual effort and domain expertise, as well as takes prompt relations into consideration for directed optimization.
Accordingly, we propose a simple, automatic and universal approach called \textsc{Prefer} (\textbf{P{\smaller R}}ompt \textbf{E}nsemble learning via \textbf{F}eedback-R\textbf{\smaller E}flect-\textbf{R}efine), towards a more effective prompt ensemble via utilizing the generative and reflective capabilities that LLMs excel at~\cite{madaan2023self}.
As shown in Figure~\ref{fig:main_idea}, our \textsc{Prefer} adopts a \textit{feedback-reflect-refine} circle for prompt boosting.
Concretely speaking, inspired by the fact that weak learners pay more attention to hard examples via weight redistribution during boosting, we propose to transfer this hard-sample-oriented weighting into nature language feedback, which returns error information to the LLM for reflection. Hence, considering the reflection information, the LLM perceives the inadequacies of existing prompts and is able to generate new prompts to refine them purposefully. Attribute to the feedback-reflect-refine path, the LLM jointly optimizes the downstream tasks solving and prompt generation in an automatic manner. 
Iterating along this path, potential conflict and redundancy among prompts are reduced, which is vital for building a more stable and faster learner.

Furthermore, to adequately unleash the ability of each prompt and further enhance the stability during boosting, we propose a bilateral bagging approach, which incorporates forward and backward thinking for multi-source verification. Specifically, drawing inspiration from human decision-making, wherein uncertain answers are often resolved through a process of elimination, we instruct the LLM to compute a confidence score for each response and subsequently filter out the most uncertain answers. Given the observed tendency of LLMs to overestimate confidence in their predictions ~\cite{zhao2021calibrate}, our bilateral bagging approach assesses the responses from both forward and backward directions, in which the overconfidence bias can be counteracted subtly. The empirical results demonstrate the superiority of our bilateral bagging approach compared to other regular methods such as majority voting in both effectiveness and efficiency.


We conduct extensive experiments and in-depth case studies on a number of tasks, including reasoning, topic classification, hate speech discrimination, etc. The empirical results testify the effectiveness of our \textsc{Prefer} approach. Moreover, \textsc{Prefer} shows superiority in both stability and efficiency compared to existing approaches. We will provide the source code for reproducibility in the supplementary material.

\section{Related Work}
Our work is conceptually related to several subareas of artificial intelligent, including Large Language Models (LLMs), prompt engineering, and prompt ensemble learning. In this section, we briefly review the works in each subarea.
\subsection{Large Language Models}
Nowadays, Large Language Models (LLMs) have made revolutionary progress and posed significant impact on various artificial intelligent community~\cite{zhao2023survey,ouyang2022training}. According to the scale law, LLMs demonstrate unprecedent power (called emergent abilities) with the rapid growth of model parameters and data volume~\cite{wei2022emergent}. For instance, the most prominent applications including ChatGPT and GPT-4~\cite{gpt4} have shown surprising reasoning ability, human-like conversation skills, as well as a rich reserve of factual commonsense.
Based on the surprising emergent abilities, a series of classical algorithms can evolve to a more intelligent version.
In this paper, we provide a pilot work on ensemble algorithm as a preliminary study. We believe that our proposed approach could not only simply serve as a strong baseline to foster future research on prompt ensemble, but also shed light on the potential research direction towards improving classical algorithms with the power of LLMs.

\begin{figure*}[t]
\setlength{\abovecaptionskip}{3pt} 
\setlength{\belowcaptionskip}{-8pt}
\centering
\includegraphics[width=0.9\textwidth]{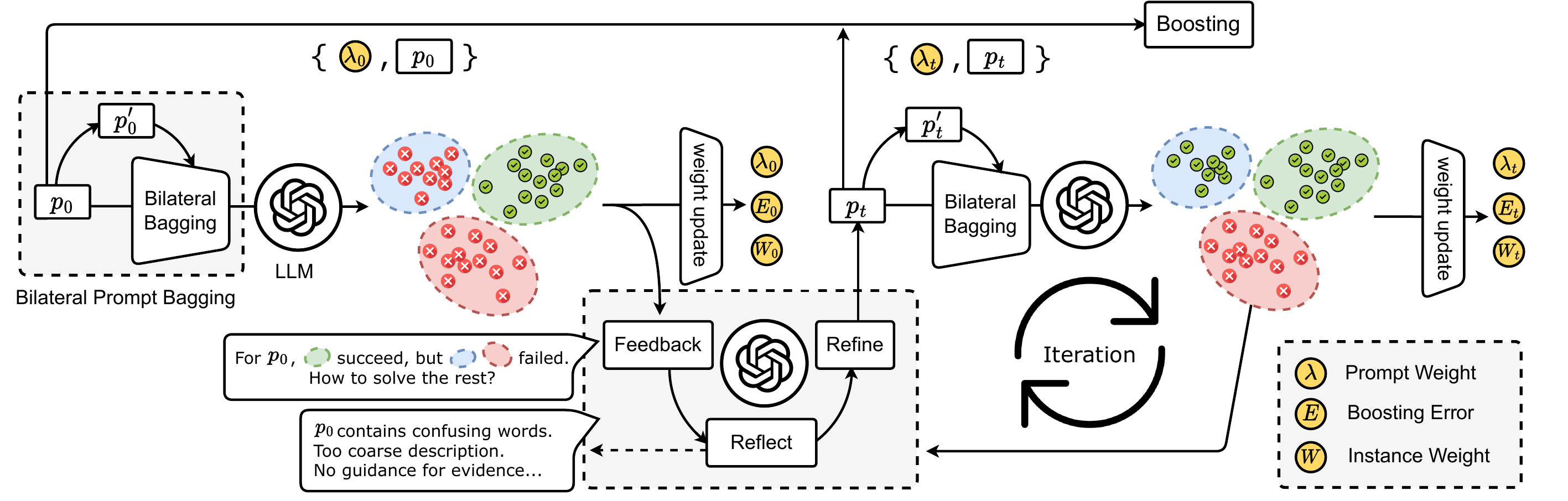} 
\caption{The pipeline of \textsc{Prefer}. Given the initial prompt $p_0$, LLM partially solves the problem via incorporating backward thinking. Then the error information will be used for prompt optimization through the feedback-reflect-refine process. Iterating this process and finally ensembling prompts based on evolved weights.}
\label{fig:pipeline}
\end{figure*}
\subsection{Prompt Engineering}
In order to invoke the power of LLMs, a series of approaches have been proposed in the community, including
parameter-efficient fine-tuning~\cite{hu2021lora,liu2021gpt} and prompt engineering~\cite{qiao2022reasoning,liu2023pre}, etc. 
Due to the heavy weight of LLMs, fully or even partly fine-tuning them is expensive and inefficient.
Accordingly, as an out-of-the-box paradigm, prompt engineering (aka prompting) has emerged as a new approach for adapting pretrain-prompt-predict path for downstream tasks. Tremendous cutting-edge effort has been made towards this area to improve the performance of prompting. 
Concretely, prompting adopts natural language as additional inputs, acting as instructions or hints to LLMs. For example, GPT2~\cite{radford2019language} allows for unsupervised learning of LLM on multiple tasks through handcrafted task-specific prompts.
However, building prompts manually can be expensive, biased and sub-optimal~\cite{liu2023pre}. Another line of works are devoted to conducting prompting in an automatic way.
STaR~\cite{zelikman2022star} utilizes a simple loop to bootstrap LLMs with a self-taught manner, in which Chain-of-Thought (CoT)~\cite{wei2022chain} rationale is iteratively generated to hint the question answering process.
Closer to our work, APO~\cite{pryzant2023automatic} iteratively optimizes the single prompt in a feedback manner, which treats the textual reflection information as gradient in classical deep learning.

\subsection{Prompt Ensemble Learning}
Prior studies have proven that LLMs have multiple  reasoning paths for a single problem, which could lead to distinct outputs from identical inputs~\cite{wang2022self}.
To this end, prompt ensemble learning has been presented as a solution, which combines several individual prompts to obtain better stability and generalization performance. 
Boosting and bagging are two typical ensemble methods widely adopted in numerous classical tasks, while their adaptation on LLMs is still in its infancy.
Current works for prompt boosting typically utilize a two-stage paradigm. 
PromptBoosting~\cite{hou2023promptboosting} has done a preliminary trial on this way, which conducts the traditional \textsc{Ada}\textsc{Boost}~\cite{freund1997decision} algorithm over a pre-defined prompt set for text classification. 
On the other hand, existing prompt bagging approaches mainly rely on regular majority voting, which can be computationally intensive. 
Notably, BPE~\cite{pitis2023boosted} focuses on constructing few-shot CoT prompts based on self-consistency~\cite{wang2022self}, which offers better performance than a single prompt in the case of introducing exponentially additional computation.
In this paper, we propose a computation-efficiency prompt bagging approach inspired by the human ethology, which incorporates prompt boosting for further performance improvement.

\section{Our \textsc{Prefer} Approach}
\subsection{Preliminaries}
In this section, we introduce preliminaries of our \textsc{Prefer} approach, including the problem formulation and the dismantling of key components.

Considering a reasoning or classification task driven by LLMs, given the training data $\mathcal{D}_{tr}=\bigcup_{i}\{(x_i,y_i)\}$, the goal of the proposed \textsc{Prefer} is to automatically construct a prompt set $\mathcal{P}=\bigcup_t\{p_t\}$ along with prompt weights $\bigcup_t\{\lambda_t\}$ via LLM-augmented ensemble learning, which can then be utilized cooperatively for the subsequent inference. Here $x_i\in\mathcal{X}$ denotes the input texts and $y_i\in{\mathcal{Y}}$ denotes the output label. It is noted that an initial prompt $p_0$ is provided as the seed for the subsequent iteration.
Instead of requiring any supervised fine-tuning (SFT) or reinforcement learning, our proposed \textsc{Prefer} utilizes out-of-box LLM API (e.g., ChatGPT or GPT-4) as the foundation model $\mathcal{M}$ for universality and flexibility.
As illustrated in Figure~\ref{fig:pipeline}, our \textsc{Prefer} mainly contains two components, i.e. feedback-driven prompt boosting and bilateral prompt bagging, which will be elaborated in sections below.

\subsection{Prompt Boosting via Feedback-Reflect-Refine} \label{sec:boosting}
Before delving into the technical details of the proposed prompt boosting approach, we first provide our design principle, based on the thinking about what characteristics should an intelligent prompt boosting have in the era of LLMs.
Review that boosting algorithms combine several individual weak learners to obtain better generalization performance. Considering the fact that weaker learners are supposed to pay more attention to hard samples during boosting, we advocate that an intelligent boosting algorithm is expected to understand what problems the previous weak learners cannot solve. That is, instead of building prompts individually, the relation among prompts should be considered for better performance and faster convergence.
In another vein, to reduce the manual effort, the prompt boosting process should be automatic, where each prompt can be constructed without manual intervention. Furthermore, the prompt boosting should be universal and adaptive, for empowering any prompting-based task with the superiority of ensemble learning seamlessly.

Our proposed \textsc{Prefer} embraces all the above design principles, towards a simple, automatic and adaptive prompt ensemble paradigm.
Inspired by the classical boosting algorithm such as \textsc{Ada}\textsc{Boost}~\cite{freund1997decision} and iterative prompting algorithms~\cite{pryzant2023automatic}, we adopt an iterative manner to build the prompt set where each prompt is treated as a weak learner.
As illustrated in Figure~\ref{fig:pipeline}, acting as a weak learner, each prompt can only handle part of the instance space, where new prompts will be added to expand the solving space by introducing more information.
Based on the error-ambiguity decomposition of ensemble learning~\cite{opitz1995generating}, the ensemble error mathematically contains two parts:
\begin{equation}
E_{ensemble}=\bar{E}-\bar{A}
\label{eq:ensemble_error}
\end{equation}
where $\bar{E}$ and $\bar{A}$ respectively denote the average error and the average ambiguity (also called diversity) of individual weak learners. Based on Eq.(\ref{eq:ensemble_error}), the ensemble performance is positively correlated with both the accuracy and diversity of weak learners.
Considering this requirement, the prompt in each iteration is supposed to focus on the hard examples that the prompts in previous iterations cannot handle. Inspired by the way human reflect and refine for improving performance when tackling difficult tasks, we propose a feedback-reflect-refine pipeline, asking the LLM to consider the relation of prompts in the iteration, generate new informative prompts, and optimize them jointly.








\begin{listing}[tb]%
\caption{{\tt solving prompt}}%
\label{lst:solving_prompt_template}%
\begin{lstlisting}[language=,breaklines=true,numbers=none,xleftmargin=0pt,xrightmargin=0pt]
# Task
Given two sentences, determine whether
sentence 2 provides an answer to the
question posed by sentence 1.

# Output format
Explain your reasoning process in one
sentence and Answer "Yes" or "No" as the
label.

# Prediction
Sentence 1: {text1}
Sentence 2: {text2}
Label:[]
\end{lstlisting}
\end{listing}

\begin{listing}[tb]%
\caption{{\tt feedback prompt}}%
\label{lst:feedback_prompt_template}%
\begin{lstlisting}[language=,breaklines=true,numbers=none,xleftmargin=0pt,xrightmargin=0pt]
I'm trying to write a Textual Entailment
task prompt. My current prompt is: {prompt}
But this prompt gets the following examples
wrong: {error_info}

Give {num_feedbacks} reasons why the prompt
could have gotten these examples wrong. Wrap
each reason with <START> and <END>.
\end{lstlisting}
\end{listing}

Concretely speaking, we define two types of prompt templates, namely the {\tt solving prompt} and the {\tt feedback prompt}, which are respectively responsible for solving downstream tasks and conducting the feedback process. Following In-Context Learning (ICL)~\cite{dai2022can}, we format both types of prompts with the component of the instruction, demonstration and output format. Exemplary cases of these two templates are illustrated in Listing~\ref{lst:solving_prompt_template} and Listing~\ref{lst:feedback_prompt_template}, respectively.
Given the initial seed prompt $p_0$ and the corresponding performance, we build the feedback prompt based on the feedback template and the wrong examples. This is reminiscent of the gradient in deep learning optimization, which indicates the direction of model optimization, the key difference lies that the feedback form changes from numerical into textual.
The feedback prompt will then be fed to the LLM $\mathcal{M}$ for self-reflecting, and $\mathcal{M}$ provides a series of reasons why the current prompt $p_t$ can solve some examples well but not others.
Based on the reflection, the LLM is asked to generate new prompts in connection with hard examples specified in the previous iteration.
In detail, the sampled wrong examples and corresponding textual labels are combined to \texttt{error\_info} in Listing~\ref{lst:feedback_prompt_template}.
Mathematically, this feedback-reflect-refine process can be formulated via the Bayesian theory:
\begin{equation}
\mathcal{P}(p_t|\mathcal{X},\mathcal{Y},p_{t-1})=\mathcal{P}(\mathcal{R}_t|\mathcal{X},\mathcal{Y},p_{t-1})\cdot\mathcal{P}(p_t|\mathcal{R}_t)
\label{bayesian}
\end{equation}
here $\mathcal{R}_t$ denotes the reflection of the LLM $\mathcal{M}$ at the $t$-th iteration.
It is noted that our \textsc{Prefer} only modifies the instruction of the {\tt solving prompt}, while other parts remain unchanged.

Close to our work, APO~\cite{pryzant2023automatic} also conducts a feedback-based mechanism for prompt optimization. Nevertheless, there are several intrinsic differences between such iterative prompting approach and our \textsc{Prefer}. 
First, APO aims to search for a single prompt covering the largest possible solution space, while our \textsc{Prefer} organizes a set of prompts via ensemble learning, which works in tandem to cover multiple sub-spaces. Second, our \textsc{Prefer} proposes an effective bagging approach to reduce the variance of the LLM, which is superior to the regular techniques such as beam search or Monte Carlo search in APO.
Experimental results demonstrate that our \textsc{Prefer} outperforms APO by a quite large margin with less computational cost and higher stability.


\subsection{Bilateral Prompt Bagging}
As shown in Eq.(\ref{eq:ensemble_error}), the quality and stability of weak learners is essential to the ensemble performance. Due to the generative property of language model, LLMs' outputs are highly sensitive to the input prompts, which affects the stability of both the feedback and weight calculation process.
To alleviate this issue, direct solutions include majority voting or beam search, which is commonly used in the community~\cite{wang2022self,li2023making}. However, these methods are computationally intensive, especially for LLMs with massive parameters.
Accordingly, to enhance the ability and stability of each prompt with limited calculation burden, we further propose a bagging approach called \textit{bilateral prompt bagging}, which draws inspiration from human behavior of utilizing forward and backward thinking for tackling difficult tasks.

\begin{algorithm}[tb]
\caption{Our \textsc{Prefer} Algorithm}
\label{alg:algorithm}
\textbf{Input}: Training data $\mathcal{D}_{tr}=\bigcup_{i}\{(x_i,y_i)\}$, the LLM $\mathcal{M}$, the seed prompt $p_0$, the prompt templates $\mathcal{T}_{\tt solving}$ and $\mathcal{T}_{\tt feedback}$\\
\textbf{Output}: the result prompt set $\mathcal{P}=\bigcup_t\{p_t\}$ and their weights $\bigcup_t\{\lambda_t\}$, the reflection set $\bigcup_t\{\mathcal{R}_t\}$
\begin{algorithmic}[1] 
\STATE Set the initial data weight to $\omega_i^{(0)}=1/|\mathcal{D}_{tr}|,\forall{i}\in\{0,\cdots,|\mathcal{D}_{tr}|\}$, $\mathcal{P}=\{p_0\}$.
\FOR{$t=0$ to $N$}
\IF{$t>0$}
\STATE Generate new $p_t$ with \{$\mathcal{M}$, reflection $\mathcal{R}_{t-1}$\}
\ENDIF
\STATE Solve target tasks with \{$p_t$, $\mathcal{T}_{\tt solving}$, $\omega_i$\}
\STATE Conduct bilateral bagging
\STATE Build \texttt{feedback prompt} with \{\texttt{error\_info}, $\mathcal{T}_{\tt feedback}$\}
\STATE Perform feedback and get the reflection $\mathcal{R}_{t}$
\STATE Compute weighted error as Eq.(\ref{eq:error_calculate})
\STATE Update the weight on $p_t$ by Eq.(\ref{eq:prompt_weight_calculate})
\STATE Update the instance weights in $\mathcal{D}_{tr}$ by Eq.(\ref{eq:dataset_weight_calculate}) followed by re-normalization
\STATE $\mathcal{P}=\mathcal{P}\cup{p_t}$, $\mathcal{R}=\mathcal{R}\cup{\mathcal{R}_t}$
\ENDFOR
\STATE \textbf{return} $\bigcup_t\{p_t\}$, $\bigcup_t\{\lambda_t\}$, $\bigcup_t\{\mathcal{R}_t\}$
\end{algorithmic}
\end{algorithm}

Concretely speaking, humans commonly adopt the process of elimination when they are not sure about the decision making. Inspired by this, we advocate that similar spirits can be utilized in the prompt bagging. In each iteration, the LLM $\mathcal{M}$ is required to evaluate its answer's confidence by utilizing the generated prompt $p_t$ followed by a confidence evaluation clause. When the evaluation result is not confident enough, the reverse thinking takes effect via conducting elimination process. In detail, we consider the quantitative confidence score evaluation in both forward and backward thinking. Take the classification task as an example, in the forward evaluation, $\mathcal{M}$ is required to measure the confidence that each candidate answer is the correct one. As for the backward evaluation, $\mathcal{M}$ is required reversely to measure the confidence that each candidate answer is excluded. For notational simplicity, we name the confidence scores corresponding to the forward and backward evaluations with $S^{+}$ and $S^{-}$ respectively. After these, the final probability can be calculated via combining $S^{+}$ and $S^{-}$ with a subtractive fashion:
\begin{equation}
\hat{y}={\arg\max}_j{\frac{e^{S^{+}_j-S^{-}_j}}{\sum_c^{K}{e^{S^{+}_c-S^{-}_c}}}}
\label{eq:explicit_bagging}
\end{equation}
here $\hat{y}$ denotes the predicted answer, $c$ and $j$ denote the indexes of candidate answers. It is noted that LLMs tend to evaluate confidence score overconfidently~\cite{zhao2021calibrate}, while our proposal ingeniously circumvents this inadequacy via positive and negative offsets. We believe that such paradigm can also shed light on the community of LLMs'
calibration~\cite{zhao2023automatic}.

Attributed to the introduction of reverse thinking mechanism, the accuracy-versus-efficiency dilemma can be largely alleviated for prompt bagging.
Experimental results explicitly manifest that such bilateral bagging outperforms regular methods (e.g., majority voting) in both effectiveness and efficiency.
\begin{table*}[t]
\setlength{\abovecaptionskip}{5pt} 
\setlength{\belowcaptionskip}{-10pt}
\centering
\begin{tabular}{l|ccccccc}
\toprule
Datasets & SNLI & MNLI  &  QNLI & RTE & Ethos & Liar & ArSarcasm \\
\midrule
Single Prompt       & 0.587 & 0.660 & 0.660 & 0.720 & 0.833 & 0.535 & 0.511 \\
Single Prompt (CoT) & 0.575 & 0.685 & 0.660 & \underline{0.731} & 0.804 & 0.549 & 0.525 \\
Synonym Ensemble    & 0.580 & \underline{0.746} & \underline{0.720} & 0.659 & 0.812 & 0.572 & 0.569 \\
PromptBoosting      & \underline{0.619} & 0.574 & 0.631 & 0.673 & -     & -     & - \\
APO                 & -      & -      & -      & -      & 0.964 & 0.663 & 0.873 \\
APO*                & -      & -      & -      & -      & \underline{0.947} & \underline{0.658} & \underline{0.639} \\
\hline
Ours             & \textbf{0.647} & \textbf{0.767} & \textbf{0.793} & \textbf{0.753} & \textbf{0.963} & \textbf{0.744} & \textbf{0.739} \\
\bottomrule
\end{tabular}
\caption{Main experimental results of our \textsc{Prefer} and the compared approaches. APO and APO* respectively denote the reported and our reproduced results of the Automatic Prompt Optimization~\cite{pryzant2023automatic}. \textbf{Bold}: best; \underline{underline}: runner-up (results are based on our reproduction).}
\label{tab:main_results}
\end{table*}
\begin{table}[t]
\setlength{\abovecaptionskip}{5pt} 
\setlength{\belowcaptionskip}{-10pt}
\centering
\begin{tabular}{l|ccc|c}
\toprule
Method & $-$Feedback & $-$Bagging & Voting & Ours \\
\midrule
SNLI    & 0.580$\downarrow$   & 0.640$\phantom{\downarrow}$ & 0.626$\phantom{\downarrow}$   & 0.647 \\
MNLI    & 0.746$\phantom{\downarrow}$   & 0.713$\phantom{\downarrow}$ & 0.733$\phantom{\downarrow}$   & 0.767 \\
QNLI    & 0.720$\phantom{\downarrow}$   & 0.747$\phantom{\downarrow}$ & 0.767$\phantom{\downarrow}$   & 0.793 \\
RTE     & 0.659$\downarrow$   & 0.740$\phantom{\downarrow}$ & 0.760$\phantom{\downarrow}$   & 0.753 \\
Ethos   & 0.812$\downarrow$   & 0.947$\phantom{\downarrow}$ & 0.938$\phantom{\downarrow}$   & 0.963 \\
Liar    & 0.572$\downarrow$   & 0.718$\phantom{\downarrow}$ & 0.701$\phantom{\downarrow}$   & 0.744 \\
Sarcasm & 0.572$\downarrow$   & 0.653$\downarrow$ & 0.649$\downarrow$   & 0.739 \\
\bottomrule
\end{tabular}
\caption{Experimental results of the ablation study. $\downarrow$ indicates a severe performance drop (more than 10\%).}
\label{tab:ablation_study}
\end{table}
\subsubsection{Overall Algorithm}
To sum up, we conclude the proposed \textsc{Prefer} in Algorithm~\ref{alg:algorithm}. Basically, our \textsc{Prefer} follows the pipeline of the classical \textsc{Ada}\textsc{Boost}~\cite{freund1997decision} algorithm, while enhancing it with the \textit{feedback-reflect-refine boosting} and the \textit{bilateral prompt bagging}. Both branches can co-adapt and cooperate for automatic prompt set optimization.
In detail, the weighted ensemble error in the $t$-th iteration is calculated as:
\begin{equation}
error^{(t)}=\sum_{i=1}^{|\mathcal{D}_{tr}|}\frac{\omega_i^{(t)}\cdot\mathbb{I}\big(y_i\neq \mathcal{M}(p_t,x_i)\big)}{\sum_i^{|\mathcal{D}_{tr}|}\omega_i}
\label{eq:error_calculate}
\end{equation}
here $\mathbb{I}$ is the identify function. Moreover, the weight in each iteration is updated based on the above error information as:
\begin{equation}
\lambda^{(t)}=\log\frac{1-error^{(t)}}{error^{(t)}}+\log\big(|\mathcal{Y}|-1\big)
\label{eq:prompt_weight_calculate}
\end{equation}
Finally, the instance weights in training dataset $\mathcal{D}_{tr}$ can be updated by:
\begin{equation}
\omega^{(t)}_i=\omega^{(t-1)}_i\cdot\exp\Big(\lambda^{(t)}\cdot\mathbb{I}\big(y_i\neq \mathcal{M}(p_t,x_i)\big)\Big)
\label{eq:dataset_weight_calculate}
\end{equation}
here $\forall{i}\in\{0,\cdots,|\mathcal{D}_{tr}|\}$ is the index of training examples.
Once the process of Algorithm~\ref{alg:algorithm} is complete, optimized prompts $\bigcup_t\{p_t\}$ along with their weights $\bigcup_t\{\lambda_t\}$ can be obtained, which can then be utilized for application via weighted decision making. Moreover, the intermediate reflection $\bigcup_t\{\mathcal{R}_t\}$ naturally provides abundant interpretability for prompt boosting.

\section{Experiments}
\subsection{Experimental Settings}
\subsubsection{Datasets}
We conduct experiments on a wide range of tasks including natural language inference and classification:
\begin{itemize}
\item Natural Language Inference\\
\textit{SNLI}~\cite{bowman2015large}, \textit{MNLI}~\cite{williams2017broad}, and \textit{RTE}~\cite{dagan2005pascal}: textual entailment inference;\\
\textit{QNLI}~\cite{rajpurkar2016squad}: question-answering inference.
\item Natural Language Classification\\
\textit{Ethos}~\cite{mollas2020ethos}: hate speech detection; 
\textit{Liar}~\cite{wang2017liar}: fake news classification; \\
\textit{ArSarcasm}~\cite{farha2020arabic}: Arabic sarcasm detection.
\end{itemize}




\subsubsection{Compared Baselines}
To manifest the superiority of our \textsc{Prefer} approach, we compare it with several state-of-the-art baselines. 
As the closest work to our proposal, PromptBoosting~\cite{hou2023promptboosting} conducts the traditional \textsc{Ada}\textsc{Boost} algorithm over a pre-defined prompt set for text classification. 
As a remarkable work of iterative prompting methods, APO~\cite{pryzant2023automatic} utilizes an iterative manner for optimizing a single prompt, where the performance of the previous prompt will be used to form a natural language ``gradient" that guides the prompt optimization.
Moreover, we also conduct single-prompt and Chain-of-Thought (CoT) enhanced single-prompt experiments, to figure out the superiority of our \textsc{Prefer} compared with vanilla and optimized non-iterative prompting works.
Lastly, we compare a variant of our \textsc{Prefer}, which rewrites synonymous prompts for boosting instead of feedback-reflect-refine paradigm, for ascertaining the utility of LLMs' reflective ability.

\subsubsection{Running settings}
To make a fair comparison, we closely follow the experimental protocols that were set up in APO with our own data split. In detail, we mainly conduct developing and evaluation of our \textsc{Prefer} in few-shot settings. For each task, we randomly sample $k$ examples from the original training dataset, to build $k$-shot training set $\mathcal{D}_{tr}$. By default, the $k$ in this paper is set to $50$. We use F1-score for performance evaluation.

\subsection{Experimental Results}
In view of the key proposals in our \textsc{Prefer} approach, we are naturally motivated to ask the following interesting research questions.
\begin{itemize}
\item \textbf{RQ1}. Is the prompt ensemble learning really useful for improving LLMs' performance?
\item \textbf{RQ2}. Are the feedback-driven boosting and bilateral bagging mechanism both useful for prompt synthesis in ensemble learning?
\item \textbf{RQ3}. Is the reason why our proposal is superior to the iterative approaches due to the expansion of the sample space?
\end{itemize}

To figure out the answers to these questions, we conduct sufficient experiments and the experimental results can be found in Table~\ref{tab:main_results}.
For the first question, we compare the ensemble-based approaches (including PromptBoosting and our \textsc{Prefer}) with the single-prompt-based approaches. As shown in the experimental results, when compared to the vanilla (Line 1) and CoT-enhanced single prompt approach (Line 2), both PromptBoosting and our \textsc{Prefer} outperform them by a significant margin.
For example, our \textsc{Prefer} outperforms the second best approach by up to 6.3\% for the \textit{QNLI} dataset, and 13.1\% for the \textit{Liar} dataset.
The general trend that becomes apparent from the results in Table~\ref{tab:main_results} is that the more difficult the task is, the better ensemble learning performs. We conjecture that it is due to the feedback-reflect-refine paradigm can achieve greater improvement for the harder tasks, while the marginal gain of this mechanism would be diminishing for easier tasks.
It is noted that the experimental results change marginally by adding Chain-of-Thought (CoT) for single-prompt approach.

To explore the second research question, we compare our \textsc{Prefer} with both the two-stage ensemble approach PromptBoosting (Line 4) and the synonym rewriting ensemble approach (Line 3). 
For PromptBoosting, we use the publicly available code of~\cite{hou2023promptboosting} and conduct experiments following its hyperparameter setting. 
For the synonym rewriting ensemble, we conduct prompt rewriting operation with same semantics, followed by regular ensemble learning similar to our \textsc{Prefer}.
As demonstrated in Table~\ref{tab:main_results}, our approach consistently outperforms the two ensemble approaches by a significant margin, reaching around 5\% to 35\% relative improvement in most datasets.
We attribute the superiority of \textsc{Prefer} to its feedback-reflect-refine mechanism as well as the design of the joint optimization paradigm that naturally captures relations among weak learners.

As for the third question, APO~\cite{pryzant2023automatic} is introduced as the remarkable approach of iterative prompting for comparison. It is noted that we reproduce the APO approach (APO* at Line 6) for a strictly fair comparison, which eliminates the interference from data sampling. Similar performance trends are observed in this comparison, that is, our \textsc{Prefer} outperforms APO with the power of feedback-reflect-refine boosting and bilateral prompt bagging. It manifests that through expanding the sample space in a nonlinear way, prompting performance can be enhanced significantly than single-prompt methods with similar iteration rounds. In fact, attributed to our bagging design, our \textsc{Prefer} is superior to APO not only in effectiveness, but also in stability and efficiency.

\subsection{Ablation Study}
\begin{figure}[t]
\centering
\includegraphics[width=0.65\linewidth]{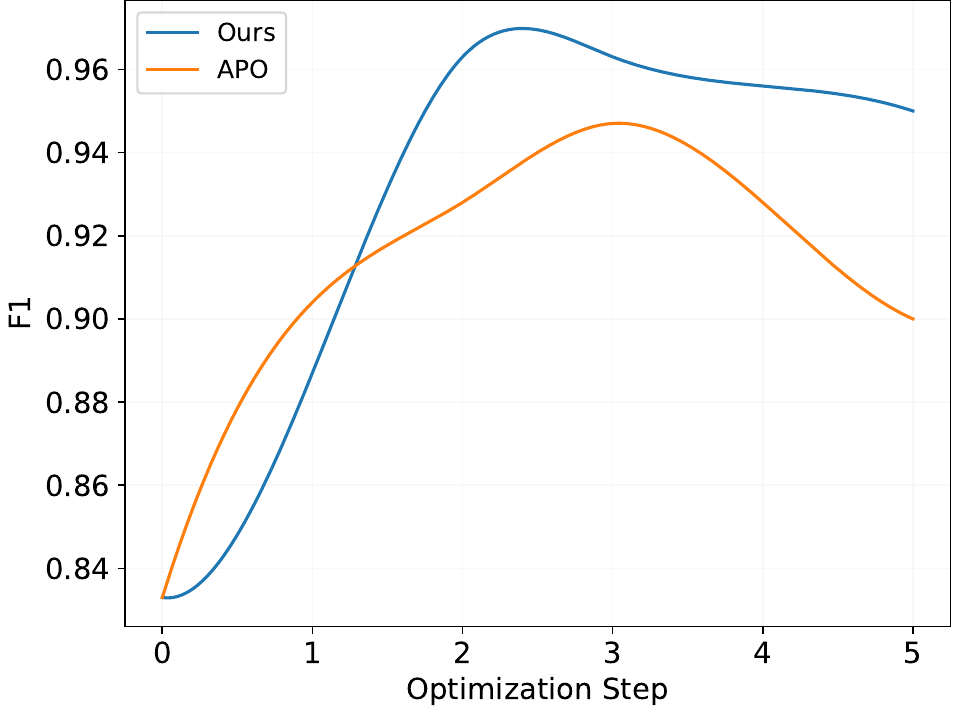} 
\vspace{-0.3cm}
\setlength{\abovecaptionskip}{10pt} 
\setlength{\belowcaptionskip}{-10pt} 
\caption{Training process comparison for APO and ours.}
\label{fig:eff_curve}
\end{figure}
To figure out the effectiveness of each component in our proposal, we perform ablations on both feedback-reflect-refine boosting and bilateral bagging, and the experimental results are provided in Table~\ref{tab:ablation_study}.
First, we remove the feedback mechanism in prompt boosting (``$-$Feedback''), in which the initial seed prompt is just modified by the LLM without directed optimization, then the similar boosting and bagging strategy is performed to align the settings of our \textsc{Prefer}.
As shown in Table~\ref{tab:ablation_study}, it is observed that the prompt ensemble without feedback-reflect-refine path is sub-optimal, signifying that such feedback mechanism plays an important role for directed prompt boosting. 
Second, to figure out the effectiveness of our bilateral bagging component, we also turn off the whole component (``$-$Bagging'') or replace it with majority voting (``Voting''), as shown in the column 3 and 4 in Table~\ref{tab:ablation_study}, respectively.
The experimental results convey that our bilateral bagging is beneficial for \textsc{Prefer}, and distinctly outperform the regular bagging approach of majority voting.
Notably, the performance of majority voting is basically satisfactory, manifesting that the prompt bagging can benefit the boosting prompt process consistently.
An interesting phenomenon is that removing the feedback-reflect-refine module leads to more serious performance decline than removing the bagging module.
This is expected, since the bagging mainly benefits the stability for each prompt, while the boosting is more important for prompt ensemble.


\subsection{Training Efficiency}
To further demonstrate the superiority of our method, we conduct detailed experiments on the \textit{Ethos} dataset for training efficiency, including training time and convergence speed.
As shown in Figure \ref{fig:eff_curve}, both APO and our \textsc{Prefer} reach the peak at optimization step 2 to 3, which indicates that neither approaches require extensive iterations to achieve impressive results. Clearly, our \textsc{Prefer} has a more stable performance retention compared to APO during subsequent iterations.
On the other hand, considering the limitations on the speed and frequency of LLM API accesses, we compare the number of API accesses during training and the time consumption for the first two prompt optimization steps, which is displayed in Table \ref{tab:eff_study}. It can be observed that the access number of APO increases rapidly during beam search and bandit selection, which brings serious efficiency problems.
On the contrary, our \textsc{Prefer} does not enforce optimal optimization at each time step, but rather maintains a stable and efficient improvement via ensemble learning.

\begin{table}[t]
\setlength{\abovecaptionskip}{5pt} 
\setlength{\belowcaptionskip}{-10pt}
\
\resizebox{0.9\linewidth}{!}{
\begin{tabular}{l|cc}
\toprule
                        & APO         & Ours   \\ 
\midrule
Frequency & $b(N+2)+T|D_{sample}|$ & $2N+2$  \\ \hline
T$_{step1}$                  & 579.0 s       & 132.4 s \\
T$_{step2}$              & 2100.4 s      & 336.1 s \\ \bottomrule
\end{tabular}
}
\caption{Comparison of training efficiency. Frequency denotes the number of API accesses required by the method within each optimization step, where $N$ is training size and $b$, $T$, $|D_{sample}|$ are hyperparameters required by APO. T$_{step1}$ and T$_{step2}$ represent the time required for the corresponding optimization steps from the beginning, where we set $N=50$, $b=4$, $T=20$, $|D_{sample}|=16$.}
\label{tab:eff_study}
\end{table}

\subsection{Case Study}
\begin{figure}[t]
\setlength{\abovecaptionskip}{5pt} 
\setlength{\belowcaptionskip}{-5pt}
\centering
\includegraphics[width=0.43\textwidth]{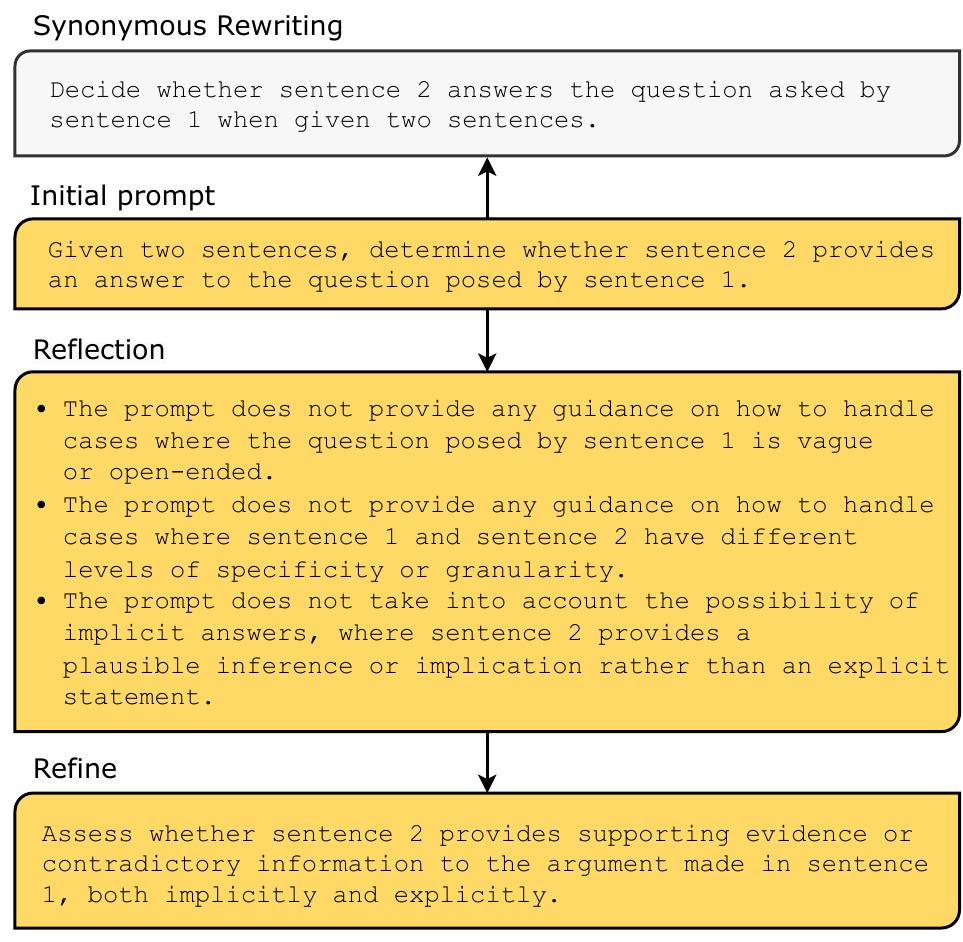} 
\caption{Comparison of the generation obtained from our feedback-reflect-refine paradigm and synonymous rewrite.}
\label{fig:case}
\vspace{-0.3cm}
\end{figure}

To visualize our feedback-reflect-refine paradigm, we provided a case study as an illustration. 
As shown in Figure \ref{fig:case}, taking the nature language inference task on the \textit{QNLI} dataset as an example, we provide the intermediate output of the LLM in the feedback-reflect-refine process, to show its effectiveness and interpretability.
Compared to the prompt generated by synonymous rewriting (gray box), the one generated by our method is more informative and logically compensates for the deficiencies of the previous prompt, thus achieving directed prompt optimization.



\section{Conclusion}
In this paper, we propose a simple, automatic, and universal prompt ensemble approach called \textsc{Prefer} (\textbf{P{\smaller R}}ompt \textbf{E}nsemble learning via \textbf{F}eedback-R\textbf{\smaller E}flect-\textbf{R}efine), empirically showing consistent and significant improvement over previous baselines.
\textsc{Prefer} contains two main components, including feedback-reflect-refine prompt boosting and bilateral prompt bagging. Prompt boosting branch directly and collectively optimizes prompt in an automatic fashion based on the evolving self-reflection. Prompt bagging proposes a bagging paradigm containing forward and backward cooperation inspired by human behavior, which adequately unearths the real quality of each generated prompt and thus ensures the stability of both the feedback-reflect-refine process and weight calculation in boosting.
In a parallel note, our \textsc{Prefer} brings the prompt ensemble approach with more interpretability by harnessing the LLMs' language ability.
For future work, two interesting questions worth studying, namely 1) how to further reduce the calculation of prompt ensemble to approach single-prompt colleagues, and 2) how to make more classical algorithm more intelligent based on the power of LLMs.

\appendix

\bibliography{aaai24}

\end{document}